\documentclass{article}
\PassOptionsToPackage{numbers, compress}{natbib}

\usepackage[preprint]{nips_2018}

\usepackage[utf8]{inputenc} 
\usepackage[T1]{fontenc}    
\usepackage{hyperref}       
\usepackage{url}            
\usepackage{booktabs}       
\usepackage{amsfonts}       
\usepackage{nicefrac}       
\usepackage{microtype}      
\usepackage{amsmath}
\usepackage{amssymb}
\usepackage[pdftex]{graphicx}
\usepackage{subcaption}
\makeatletter

\renewenvironment{thebibliography}[1]
{\section*{\refname\@mkboth{\refname}{\refname}}%
   \list{\@biblabel{\@arabic\c@enumiv}}%
        {\settowidth\labelwidth{\@biblabel{#1}}%
         \leftmargin\labelwidth
         \advance\leftmargin\labelsep
     \setlength\itemsep{0.5pt}
     \setlength\baselineskip{9.2pt}
         \@openbib@code
         \usecounter{enumiv}%
         \let\p@enumiv\@empty
         \renewcommand\theenumiv{\@arabic\c@enumiv}}%
   \sloppy
   \clubpenalty4000
   \@clubpenalty\clubpenalty
   \widowpenalty4000%
   \sfcode`\.\@m}
  {\def\@noitemerr
    {\@latex@warning{Empty `thebibliography' environment}}%
   \endlist}

\def\JTeX{\leavevmode\lower .5ex\hbox{J}\kern-.17em\TeX}
\def\JLaTeX{\leavevmode\lower.5ex\hbox{J}\kern-.17em\LaTeX}

\makeatother 

\title{Zero-shot  Domain Adaptation without \\ Domain Semantic Descriptors}

\author{
  Atsutoshi Kumagai \\
   NTT Secure Platform Laboratories\\
  \texttt{kumagai.atsutoshi@lab.ntt.co.jp} \\
  \And
  Tomoharu Iwata \\
  NTT Communication Science Laboratories \\
  \texttt{iwata.tomoharui@lab.ntt.co.jp} \\
}

\begin{document}

\maketitle

\begin{abstract}
We propose a method to infer domain-specific models such as classifiers for unseen domains,
from which no data are given in the training phase, without domain semantic descriptors.
When training and test distributions are different, standard supervised learning methods perform poorly.
Zero-shot domain adaptation attempts to alleviate this problem by inferring models that
generalize well to unseen domains by using training data in multiple source domains.
Existing methods use observed semantic descriptors characterizing domains such as time information to infer
the domain-specific models for the unseen domains.
However, it cannot always be assumed that such metadata can be used in real-world applications. 
The proposed method can infer appropriate domain-specific models without any semantic descriptors
by introducing the concept of {\it latent domain vectors}, which are latent representations for the domains 
and are used for inferring the models.
The latent domain vector for the unseen domain is inferred from the set of the feature vectors in the corresponding domain, 
which is given in the testing phase.
The domain-specific models consist of two components:
the first is for extracting a representation of a feature vector to be predicted, 
and the second is for inferring model parameters given the latent domain vector.
The posterior distributions of the latent domain vectors and the domain-specific models are parametrized by neural networks, 
and are optimized by maximizing the variational lower bound using stochastic gradient descent.
The effectiveness of the proposed method was demonstrated through experiments using one regression and two classification tasks.
\end{abstract}

\section{Introduction}
Many supervised learning methods rely heavily on the assumption 
that training and test data follow the same distribution.
However, this assumption is often violated in real-word applications.
For example, in computer vision, images taken with different cameras or 
in different conditions follow different distributions
\cite{2011unbiased}.
In sentiment analysis, reviews on different product categories follow different distributions
\cite{blitzer2007,glorot2011}.
When the training and test distributions are different, 
standard supervised learning methods perform significantly worse
\cite{ben2007analysis,saenko2010adapting,donahue2014decaf}.

Although large labeled data drawn from the test distribution can alleviate this problem,
such data are often time-consuming and impractical to collect
since labels need to be manually assigned by domain experts.
Domain adaptation aims to learn models that perform well on a testing domain, called a target domain,
by using data in a training domain, called a source domain
~\cite{tzeng2014deep,long2016unsupervised,long2016deep,duan2009domain,saenko2010adapting}.
Learning such models requires a few labeled and/or unlabeled data in the target domain during training.
However, training after obtaining data from the target domain is problematic in some real-world applications.
For example, with the growth of the Internet of Things (IoT), 
the importance of performing prediction on devices such as 
speech recognition on mobile devices
\cite{schuster2010speech}
and character recognition on portable devices
\cite{xiao2017building}
is increasing.
Since these devices do not have sufficient computing resources,
training on these devices is difficult even if new target domains appear that contain training data.
In security, the need to protect a wide variety of devices such as sensors, cameras, and cars
from cyber attacks is rapidly rising
\cite{babar2010proposed}.
Since the new devices (target domains) appear one after another,
it is difficult to protect all these devices quickly with time-consuming training.
In personalized services such as e-mail systems,
personalization (training) for new users is allowed only when the users allow the use of their data for training to preserve privacy.

Zero-shot domain adaptation aims to adapt target domains where there are no data in the training phase
by using data in multiple source domains
\cite{yang2014unified,yang2015zero,muandet2013domain,motiian2017unified,li2017learning}.
We call these target domains {\it unseen} domains.
Since these methods do not require any data in the target domains, 
they can instantly adapt to various unseen domains without time and memory consuming training.
It is important to infer an appropriate domain-specific model for each domain in the training phase
since the characteristics of each domain differ.
Existing methods infer such models by using additional semantic descriptors characterizing domains
such as time information
\cite{lampert2015predicting,kumagai2016,kumagai2017},
device and location information
\cite{yang2014unified},
and pose and illumination information
\cite{qiu2012domain}.
Although they improve performance,
assuming semantic descriptors restricts the applicability of zero-shot domain adaptation
because semantic descriptors cannot always be obtained.
Recently, Shankar et al.
\cite{shankar2018generalizing}
proposed a method to infer domain-specific models without semantic descriptors.
This method infers the domain-specific model given {\it one} feature vector from the unseen domain in the testing phase.
However, it is difficult to infer such models correctly from {\it one} feature vector
because the domain is usually characterized by the data distribution, which requires the {\it set} of feature vectors to be estimated.

In this paper, we propose a method to infer appropriate domain-specific models for unseen domains without 
semantic descriptors given the sets of unlabeled data in the corresponding domains at the testing phase and labeled data in multiple source domains in the training phase.
Once training is executed by using labeled data in multiple source domains,
the proposed method can infer domain-specific models for the unseen domains given the sets of the feature
vectors in the corresponding domains.
Instead of using semantic descriptors, the proposed method introduces the concept of {\it latent domain vectors}, which are latent representations for domains and are used to infer the models.
Domain-specific models are modeled by two types of neural networks:
the first extracts features, i.e., it outputs a representation given an observed feature vector, and
the second outputs model parameters given the latent domain vector.
By integrating the outputs of both neural networks,
the models can make a prediction for the feature vector of each domain.
The latent domain vector is estimated from the {\it set} of the feature vectors rather than {\it one} feature vector, 
which enables us to accurately infer the models reflecting the characteristics of the domains.
To infer the latent domain vectors from the sets of the feature vectors in different domains,
the neural networks need to take sets with different sizes as inputs.
However, traditional neural networks take vectors with a fixed size as inputs and cannot deal with sets.
To overcome this problem,
we employ the architecture of deep sets
{\cite{NIPS2017_6931}, which is permutation invariant to the order of samples in the sets
and thus can take the sets as inputs.
Specifically, we model the parameters for the posterior distribution of the latent domain vector given the set of feature vectors by the permutation invariant neural networks. 
Both neural networks for the latent domain vectors and domain-specific models are simultaneously optimized by maximizing the variational lower bound using stochastic gradient descent (SGD).
Since the proposed method is based on a Bayesian framework,
it can infer models by naturally considering the uncertainty of estimated latent domain vectors,
which enables robust prediction.

\section{Proposed Method}

\subsection{Notations and Task}
We introduce the notations used in this paper and define the task we investigate.
We treat a multi-class classification problem as a running example 
though the proposed method can be applied to other supervised learning tasks.
Let ${\cal D} _d := \{ ( {\bf x} _{dn}, y_{dn} ) \} _{n=1}^{N_d} $ be a set of labeled data in the $d$-th domain, 
where ${\bf x} _{dn} \in \mathbb{R}^{M} $ is the $M$-dimensional feature vector of the $n$-th sample 
of the $d$-th domain, $y_{dn} \in \{1,\dots,C \} $ is its label, and $N_d$ is the number of the labeled data in the $d$-th domain.
${\bf X} _d := \{ {\bf x} _{dn} \}_{n=1}^{N_d}$ and ${\bf Y} _d := \{ y_{dn} \}_{n=1}^{N_d} $ represent a set of feature vectors and labels in the $d$-th domain,
respectively.
We assume that feature vector size $M$ and the number of classes $C$ are the same in all domains.
Assume we have labeled data in $D$ source domains, ${\cal D} := \bigcup  _{d=1}^{D} {\cal D} _d $ during training. 
Our goal is to predict a classifier for the $d'$-th domain where any $d' \notin \{1,\dots,D\}$, $h_{d'}: \mathbb{R}^{M} \rightarrow \{ 1, \dots, C \} $,
which can accurately classify samples in the $d'$-th domain, when the set of the samples $ {\bf X} _{d'} = \{ {\bf x} _{d'n} \} _{n=1}^{N_{d'}} $ is given in the testing phase.
Figure \ref{ov} illustrates the proposed method.
\begin{figure*}[t]
  \begin{minipage}{0.49\hsize}
     \centering
      \includegraphics[width=6.5cm]{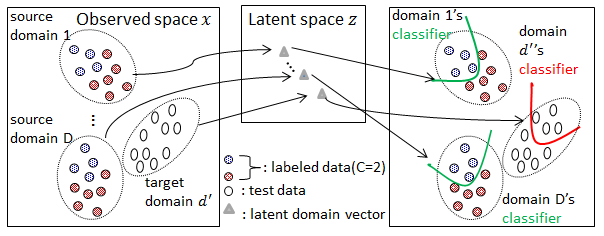}
      \caption{Illustration of the proposed method. Each domain is represented by a latent domain vector, and classifiers are inferred by using the latent domain vectors.
      After training, our method can infer classifiers for unseen domains given the sets of feature vectors in the domains.}
      \label{ov}
  ~~
  \end{minipage}
  ~
  \begin{minipage}{0.49\hsize}
      \centering
      \includegraphics[width=4.0cm]{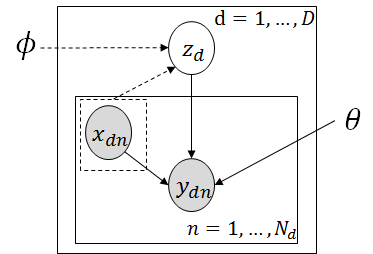}
      \caption{Graphical model representation of the proposed model. Solid lines denote our generative model, and dashed lines denote the
      inference model to approximate posterior $q_{\bf \phi} ({\bf z} _d | {\bf X} _d )$. The posterior of the domain vector ${\bf z} _{d}$ is estimated from the set of feature vectors ${\bf X} _d = \{ {\bf x} _{dn} \} _{n=1}^{N_d}$.}
      \label{gm}
  \end{minipage}
\end{figure*}

\subsection{Model}
The proposed method assumes that each domain has a $K$-dimensional latent continuous variable ${\bf z} _{d} \in \mathbb{R}^{K} $, which is called a {\it latent domain vector} in this paper. 
This latent domain vector ${\bf z} _{d}$ is generated from a standard Gaussian distribution ${\cal N} ({\bf z} _{d} | {\bf 0}, {\bf I} )$.
Each label $y_{dn}$ of a feature vector ${\bf x} _{dn}$ is generated depending on this feature vector and the corresponding latent domain vector ${\bf z} _{d}$, $p_{\bf \theta} (y_{dn}|{\bf x} _{dn}, {\bf z} _{d} ) $, which is modeled by neural networks with parameters ${\bf \theta}$.
Specifically, the $c$ th element of the output layer of $p_{\bf \theta} (y_{dn}|{\bf x} _{dn}, {\bf z} _{d} ) $ before applying the softmax function 
is modeled as
\begin{align}
\label{outputs}
f_c ( {\bf x} _{dn}, {\bf z} _{d} ) := h ({\bf x} _{dn}) \cdot g_c ({\bf z} _{d}), \ \ \ h ({\bf x} _{dn}) \in \mathbb{R}^{J}, \ g_c ({\bf z} _{d}) \in \mathbb{R}^{J}
\end{align}
where $h$ and $g_c$ are neural networks, 
$J$ is the output size of the neural networks, and $\cdot$ is an inner product.
Intuitively, $h({\bf x} _{dn})$ is a representation after feature extraction, and ${\bf g } ({\bf z} _{d}) := (g_1({\bf z} _{d}), \dots, g_C ({\bf z} _{d}) ) \in \mathbb{R}^{J \times C}$ is classifier parameters for the $d$-th domain.
By changing the latent domain vector ${\bf z} _{d}$, the proposed method can represent various class decision boundaries.
Note that this formulation is similar to the model described by Yang and Hospedales
\cite{yang2014unified}
although
${\bf z} _{d}$ represents an observed semantic descriptor characterizing a domain in their work
\cite{yang2014unified}.

The marginal log-likelihood of our model on the labeled training data ${\cal D}$ is given by
\begin{align}
\label{joint_dist_all}
{\ln} \ p({\cal D} ) = {\ln} \ \prod _{d=1}^{D}  \int \left[ \prod _{n=1}^{N_d} p_{\bf \theta} (y_{dn} |{\bf x} _{dn}, {\bf z} _{d} ) \right] p({\bf z} _{d}) d {\bf z} _{d}.
\end{align}
Since $p_{\bf \theta} (y_{dn} |{\bf x} _{dn}, {\bf z} _{d} )$ is modeled by neural networks, 
analytically obtaining the posterior of the latent domain vector is intractable.
Therefore, we approximate the posterior of the latent domain vector with $q_{\bf \phi}$,
which is modeled as
\begin{align}
\label{recog}
q_{\bf \phi} ({\bf z} _d | {\bf X} _d ) = {\cal N} ({\bf z} _d |  \mu_{\bf \phi} ({\bf X} _d), \sigma_{\bf \phi}^2 ({\bf X} _d) ),
\end{align}
where mean $\mu_{\bf \phi} ({\bf X} _d)  \in \mathbb{R}^{K} $ and variance $\sigma_{\bf \phi}^2 ({\bf X} _d)  \in \mathbb{R}^{K} $
are modeled by neural networks with parameters ${\bf \phi}$.
In this model, the latent domain vector ${\bf z} _d $ depends only on the set of feature vectors ${\bf X} _d$
although we can model so that ${\bf z} _d$ depends on both ${\bf X} _d$ and ${\bf Y} _d$.
However, by this modeling,
the proposed method can predict the latent domain vectors of unseen domains when the sets of the unlabeled data
in these domains are only given in the test phase.
Therefore, we can infer appropriate classifiers for the unseen domains without training.

Since the $q_{\bf \phi}$ deals with the set of feature vectors ${\bf X} _d$ as an input, 
the neural networks for the parameters $\mu_{\bf \phi} ({\bf X} _d)$ and $\sigma_{\bf \phi}^2 ({\bf X} _d)$ must
be permutation invariant to the order of samples in the set.
In addition, since different domains will have different numbers of samples, 
these neural networks should be robust to differences in the data size.
For neural networks satisfying these conditions,
we use the following neural network architecture,
\begin{align}
\label{invariant}
\tau ({\bf X} _d) = \rho \left( \frac{1}{N_d} \sum_{n=1}^{N_d} \eta ( {\bf x} _{dn} ) \right),
\end{align} 
where $\tau ({\bf X} _d)$ represents one of the $\mu_{\bf \phi} ({\bf X} _d)$ and ${\ln} \ \sigma_{\bf \phi}^2 ({\bf X} _d)$, $\rho$ and $\eta$ are any neural networks,
respectively. This neural network is obviously permutation invariant and
reduces the influence of the data size by simply taking the average of $\eta ( {\bf x} _{dn} )$.
Note that this architecture is a kind of permutation invariant architectures proposed by Zaheer et al.
\cite{NIPS2017_6931}.
Figure \ref{gm} shows a graphical model representation of the proposed model, where the shared and unshared nodes indicate observed and latent variables, respectively.

\subsection{Learning}
We can derive a lower bound on the marginal log-likelihood ${\ln} \ p({\cal D} )$ using $q_{\bf \phi} ({\bf z} _d | {\bf X} _d )$:
\begin{align}
\label{elbo}
{\rm ln} \ p({\cal D}) \geqq {\cal L} ({\cal D} ; {\bf \theta}, {\bf \phi}):= \sum_{d=1}^{D} \left[ - D_{KL} ( q_{\bf \phi} ({\bf z} _{d} | {\bf X} _d) \| p({\bf z} _{d}))
+ \mathbb{E} _{q_{\bf \phi} ({\bf z} _{d} | {\bf X} _d)} \left[ \sum_{n=1}^{N_d} {\rm ln} \ p_{\bf \theta} (y_{dn}|{\bf x} _{dn}, {\bf z} _{d} ) \right] \right],
\end{align}
where $D_{KL} ( q_{\bf \phi} ({\bf z} _{d} | {\bf X} _d) \| p({\bf z} _{d}))$ denotes the Kullback Leibler (KL) 
divergence between the approximate posterior $q_{\bf \phi} ({\bf z} _{d} | {\bf X} _d)$ and the prior $p({\bf z} _d)$ for the $d$-th latent domain vector. The parameters of the neural networks for the classifiers and latent domain vectors,
${\bf \theta}$ and ${\bf \phi}$, are obtained by maximizing this lower bound $ {\cal L} ({\cal D} ; {\bf \theta}, {\bf \phi})$ using SGD. 
Although the expectation term of \eqref{elbo} is still intractable, this term can be effectively approximated by 
the reparametrization trick
\cite{kingma2013auto}:
That is, we draw $L$ samples ${\bf z} _{d}^{(\ell)} \sim q_{\bf \phi} ({\bf z} _{d} | {\bf X} _d)$
by ${\bf z} _{d}^{(\ell)} = \mu_{\bf \phi} ({\bf X} _d) $ + $\epsilon^{(\ell)} \sigma_{\bf \phi} ({\bf X} _d) $,
where $\epsilon^{(\ell)} \sim {\cal N} ( {\bf 0} , {\bf I})$,
and we have
$\mathbb{E} _{q_{\bf \phi} ({\bf z} _{d} | {\bf X} _d)} [ {\rm ln} \ p_{\bf \theta} (y_{dn}|{\bf x} _{dn}, {\bf z} _{d} ) ] \approx 
\frac{1}{L} \sum_{\ell=1}^{L} {\rm ln} \ p_{\bf \theta} (y_{dn}|{\bf x} _{dn}, {\bf z} _{d}^{(\ell)} )$.
As a result, the objective function to be maximized with respect to the parameters ${\bf \theta}$ and ${\bf \phi}$ becomes
\begin{align}
\label{last_obj}
{\cal L} ({\cal D} ; {\bf \theta}, {\bf \phi} ) \approx \sum_{d=1}^{D} \left[ - D_{KL} ( q_{\bf \phi} ({\bf z} _{d} | {\bf X} _d) \| p({\bf z} _{d})) +  \frac{1}{L} \sum_{\ell=1}^{L} \sum_{n=1}^{N_d} {\rm ln} \ p_{\bf \theta} (y_{dn}|{\bf x} _{dn}, {\bf z} _{d}^{(\ell)} ) \right].
\end{align}

\subsection{Prediction}
Given the set of feature vectors from the unseen domain ${\bf X} _{d'} = \{ {\bf x} _{d'n} \} _{n=1}^{N_{d'}}$, 
the proposed method predicts the distribution of the label given the feature vector ${\bf x} _{d'n} $ as follows:
\begin{align}
\label{pred}
p(y_{d'n} | {\bf x} _{d'n}) = \int p_{\bf \theta} (y_{d'n}|{\bf x} _{d'n}, {\bf z} _{d'} ) q_{\bf \phi} ({\bf z} _{d'} | {\bf X} _{d'}) d{\bf z} _{d'}
\approx \frac{1}{L} \sum_{\ell=1}^{L} p_{\bf \theta} (y_{d'n}|{\bf x} _{d'n}, {\bf z} _{d'}^{(\ell)} ), 
\end{align}
where ${\bf z} _{d'}^{(\ell)} $ = $\mu_{\bf \phi} ({\bf X} _{d'}) $ + $\epsilon^{(\ell)} \sigma_{\bf \phi} ({\bf X} _{d'}) $ and
$\epsilon^{(\ell)} $ is a sample drawn from ${\cal N} ( {\bf 0} , {\bf I})$. 
The proposed method can predict the labels of the feature vectors while taking into account the uncertainty of the latent domain vectors 
by sampling ${\bf z} _{d'}$ from the posterior distribution $q_{\bf \phi} ({\bf z} _{d'} | {\bf X} _{d'}) $, which enables robust classification. 

\section{Related Work}
Many domain adaptation or transfer learning methods have been proposed under many different settings.
Unsupervised domain adaptation aims to adapt a target domain by using labeled source data and unlabeled target data
\cite{tzeng2014deep,long2016unsupervised,long2016deep,kon2017cvpr,nips2017benaim}.
Semi-supervised domain adaptation methods use a small amount of labeled target data as well as these data
\cite{dai2007,daume2009frustratingly,duan2009domain,saenko2010adapting}.
Some methods assume multiple source domains for adaptation
\cite{mansour2009domain,zhang2015multi}.
Some methods discover latent domains in a source domain to improve performance
\cite{hoffman2012discovering,xiong2014latent,mancini2018boosting}
All these methods assume some data in the target domain can be used for training. 
However, our task does not use any data in the target domain during training.

Multi-task learning improves performances of several tasks simultaneously by sharing training data 
obtained from multiple domains
\cite{ruder2017overview}.
These methods usually assume that data for the domains, where performance improvement is desired,
can be used during training.
In contrast, the proposed method aims to improve performance for the unseen domains, where there are no data during training.

Zero-shot learning attempts to learn classifiers that generalize to unseen classes that do not appear in the training phase
\cite{xian2017zero}.
Our task is to generalize to unseen domains in the training phase.

The term {\it zero-shot domain adaptation} has been used by Yang and Hospedales
\cite{yang2014unified,yang2015zero} and is sometimes called {\it domain generalization}.
Many zero-shot domain adaptation methods attempt to find domain-invariant representations or models for generalization.
Muandet et al.
\cite{muandet2013domain}
proposed a method to learn kernel-based representations that minimize the domain discrepancy.
Ghifary et al.
\cite{ghifary2015domain}
proposed a method to learn domain-invariant features by using multi-task auto-encoders.
Some methods assume model parameters consisting of a domain agnostic term and a domain specific term,
and the agnostic parameters are only used for unseen domains
\cite{khosla2012undoing,li2017deeper}.
Li et al.
\cite{li2017learning}
used meta-learning methods to generalize models.
Higgins et al.
\cite{higgins2017darla}
aimed to learn domain-invariant disentangled features for reinforcement learning.
Although these methods are somewhat effective, 
they ignore specific information for each domain and use only common parts of the existing domains.
In contrast, the proposed method can infer domain-specific models by using the set of
feature vectors in each domain.

Some methods assume the additional semantic descriptors can be used that represent domains 
such as time information
\cite{lampert2015predicting,kumagai2016,kumagai2017},
device and location information
\cite{yang2014unified},
and instructions for reinforcement learning
\cite{oh2017zero}.
Although these semantic descriptors enhance the generalization ability,
such data cannot be used under some circumstances.
The proposed method can be applied in these situations without any semantic descriptors.

One method, called CROSSGRAD
\cite{shankar2018generalizing},
has been recently proposed to infer domain-specific models that generalize to unseen domains
by using domain signals
without semantic descriptors.
This method utilizes domain signals similarly to the proposed method,
but there are clear differences in several aspects.
First, both methods infer domain-specific models differently.
This method infers the model from {\it one} feature vector.
Since the same model is inferred as long as the same feature vector is given, 
this model cannot output different labels given the same feature vector.
This fact restricts the generalization capability of this method.
For example, this method cannot cope with the situation in which
the distributions of labels given feature vectors $p(y|{\bf x})$ differ between domains.
In contrast, the proposed method can deal with this since
the model is inferred from the {\it set} of feature vectors and
thus different models can be inferred between different domains.
Second, this method does not take into account the uncertainty of the estimated latent domain vector 
even though this estimation is not easy and the latent domain vectors contribute greatly to performance.
In contrast, the proposed method naturally takes into account such uncertainty owing to a Bayesian framework.
Third, this method uses data augmentation techniques based on adversarial training to avoid over-fitting to the source domains.
However, generally, such adversarial training is unstable and often leads to poor performance
\cite{salimans2016improved}.
The proposed method can generalize to the unseen domains without such adversarial training as 
shown in our experiments.

\section{Experiments}
In this section, we demonstrate that the proposed method has better zero-shot domain adaptation ability than 
three different comparison methods in two classification tasks and one regression task.
To measure this ability, we evaluated the accuracy for the classification tasks and Root Mean Square Error (RMSE)
for the regression task on target domains that were not used for training.

\subsection{Datasets}
We used three real-world datasets: MNIST-r\footnote{https://github.com/ghif/mtae} and IoT\footnote{https://archive.ics.uci.edu/ml/datasets/detection\_of\_IoT\_botnet\_attacks\_N\_BaIoT} for
classification and School
\footnote{http://multilevel.ioe.ac.uk/intro/datasets.html} for regression.

The MNIST-r dataset is commonly used in zero-shot domain adaptation studies.
This dataset, which was derived from the hand written digit dataset MNIST, was introduced by Ghifary et al.
\cite{ghifary2015domain}.
Each domain is created by rotating the images in multiples of 15 degrees: 0,15,30,45,60, and 75. 
Therefore, this dataset has six different domains. 
Each domain has 1,000 images, which are represented by 256-dimensional vectors, of 10 classes (digits).
We tested on one domain while training on the rest by changing the test domain.
In each domain used for training, we randomly selected 80\% of samples for training and 20\% of samples
for validation. 

The IoT dataset contains real network traffic data, which are gathered from nine IoT devices infected by BASHLITE malware.
This dataset has normal and malicious traffic data, which are divided into five attacks.
Each data point is represented by a 115-dimensional vector and the task is to classify each data point as normal or malicious.
We normalized each feature vector by ${\ell} _2$-normalization.
We did not use the devices that had no normal data.
We treated each device as a domain and tested on one domain while training on the rest by changing the test domain.
We randomly choose samples from the first 5,000 traffic data from each normal/attack file of each device file
while equalizing the ratio of normal and malicious data.
As a result, for each trial, we used 14,000 samples for training, 1,400 samples for validation, and 3,800 samples for testing.

The School dataset is well used in multi-task learning studies.
This dataset consists of exam grades of 15,362 students from 139 schools.
Given the 28 features representing a student, the task is to predict each student's exam grades.
By regarding each school as a domain, this dataset has 139 different domains. 
In our experiments, we used four datasets with different numbers of source domains.
Specifically, we randomly choose 10\%, 15\%, 25\%, and 50\% of domains for training and
the remaining 90\%, 85\%, 75\%, and 50\% of domains for testing.
Then, in each domain used for training, we randomly selected 80\% of samples for training and 20\% of samples
for validation. 

\subsection{Comparison Methods}
We compared the proposed method with two popular zero-shot domain adaptation methods, 
classification and contrastive semantic alignment (CCSA)
\cite{motiian2017unified}
and CROSSGRAD
\cite{shankar2018generalizing},
and one baseline method.
None of methods, including the proposed method, uses any semantic descriptors.

Baseline is a neural network, where no adaptation is performed and is learned by all training data
${\cal D}$ without identification of domains. 
For MNIST-r, we used a neural network with one dense hidden layer with 100 hidden nodes, ReLU 
activations, and a softmax output function.
For IoT, we used a neural network with one dense hidden layer with 50 hidden nodes, ReLU 
activations, and a softmax output function.
For School, we used a neural network with one dense hidden layer with 10 hidden nodes and ReLU
activations.

CCSA is widely used zero-shot domain adaptation method that finds domain-invariant representations.
This method uses the contrastive semantic alignment loss, which 
brings samples with the same labels closer and separates samples with different labels in the embedding spaces,
to learn representations.
This loss is added to the hidden layers of the baseline network.

CROSSGRAD is a zero-shot domain adaptation method that
infers the domain-specific model from {\it one} feature vector. 
In CROSSGRAD, the domain features are incorporated in the baseline network by concatenating with the inputs the same as
\cite{shankar2018generalizing}.
The neural network for a domain classifier is the same architecture as the baseline network for MNIST-r and IoT except for 
the number of output nodes.
For School, we used a neural network with two dense hidden layers with 10 hidden nodes and ReLU activations
for the domain classifier since it improved performance.

For the proposed method, to infer the latent domain vectors, we used neural network architectures 
which have the same numbers of hidden layers and nodes and the same activations as that for the domain classifier in CROSSGRAD. 
Specifically, for $\mu_{\bf \phi} ({\bf X} _d)$ and ${\ln} \ \sigma_{\bf \phi}^2 ({\bf X} _d)$ of  $q_{\bf \phi} ({\bf z} _d | {\bf X} _d )$, 
the shared single (two)-layer neural networks are used as the neural network $\eta$ in \eqref{invariant} for MNIST-r and IoT (School),
and different single-layer neural networks are used as $\rho$ in \eqref{invariant}.
For $p_{\bf \theta} (y_{dn}|{\bf x} _{dn}, {\bf z} _{d} ) $,
$h({\bf x} _{dn})$ in \eqref{outputs} corresponds to the activated hidden layer
of the baseline network, 
and $g_c$ in \eqref{outputs} is the single-layer neural network.
To prevent the outputs of $g_c$ becoming too large since it leads to unstable performance, 
tanh activations are applied to the outputs, which improved the performance. 
The proposed method was implemented by using a deep learning framework, Chainer
\cite{tokui2015chainer}.

\subsection{Hyper-parameter Settings}
For all experiments, we selected hyper-parameters on the basis of accuracy for MNIST-r and IoT and RMSE for School 
on the validation data. For CCSA, the regularization parameter $\gamma$ was chosen from $\{0.01,0.1,0.25,0.5\}$, 
and the number of samples for the contrastive semantic alignment loss was set to two for all datasets.
For CROSSGRAD, the data augmentation weight $\alpha = \alpha_l = \alpha_d$ was chosen from $\{ 0.1,0.25,0.5,0.75,0.9\}$, 
and the step size for data augmentation $\epsilon$ was selected from $\{ 0.5,1.0,2.0,2.5\}$ the same as
\cite{shankar2018generalizing}.
For the proposed method, the dimension of the latent domain vector $K$ was chosen from $\{2,3,\dots,10\}$, 
and the sample size of the reparametrization trick $L$ was set to one for training and ten for testing. 
For all methods, we used the Adam optimizer
\cite{kingma2014adam}
with a learning rate of $0.001$. The maximum number of epochs was 300 for MNIST-r and School and 200 for IoT, 
and we evaluated the results when they got the best accuracies for MNIST-r and IoT,
and best RMSEs for School on the validation data after 15 epochs to avoid over-fitting.
For the proposed method, the minibatch size should preferably be somewhat large
since the latent domain vectors are estimated from the {\it sets} of the feature vectors.
Thus, we set the minibatch size as 512 for all methods.
We conducted experiments on ten randomized trials for each test domain, and reported the mean accuracy for MNIST-r and IoT and mean RMSE for School.
\begin{table*}[t]
\caption{Average and standard deviation of accuracies with different test domains with MNIST-r and IoT, and 
average and standard deviation of RMSE over different test domains with School.
Values in boldface are statistically better than others (in paired t-test, $p=0.05$).
The bottom row gives the number of best cases of each method.}
\label{table:t1}
\centering
\begin{tabular}{llcccc}
\hline
\multicolumn{1}{c}{Data} & \multicolumn{1}{c}{Test domain} & \multicolumn{1}{c}{Proposed} & \multicolumn{1}{c}{Baseline} & \multicolumn{1}{c}{CCSA} & \multicolumn{1}{c}{CROSSGRAD} \\
\hline
MNIST-r & 0 deg. & \textbf{77.44$\pm$2.01} & 71.63$\pm$1.30 & 70.36$\pm$1.35 & 74.71$\pm$0.79 \\
 & 15 deg. & \textbf{94.29$\pm$0.75} & 91.46$\pm$5.30 & 91.47$\pm$0.75 & 91.96$\pm$0.74 \\
 & 30 deg. & \textbf{91.17$\pm$0.89} & 88.44$\pm$1.11 & 87.19$\pm$0.90 & 88.09$\pm$0.67 \\
 & 45 deg. & \textbf{77.46$\pm$1.60} & 75.21$\pm$1.22 & 74.03$\pm$1.39 & 72.84$\pm$1.08 \\
 & 60 deg. & \textbf{92.12$\pm$0.71} & 89.02$\pm$0.87 & 87.35$\pm$0.76 & 88.89$\pm$0.55 \\
 & 75 deg. & \textbf{80.09$\pm$1.46} & 75.71$\pm$0.97 & 75.44$\pm$0.70 & 76.51$\pm$1.24 \\
\hline
IoT & D-Doorbell & \textbf{99.27$\pm$0.09} & 95.86$\pm$0.33 & 95.44$\pm$0.49 & 95.54$\pm$0.73 \\
 & Thermostat & \textbf{99.35$\pm$0.08} & \textbf{99.38$\pm$0.12} & \textbf{99.35$\pm$0.16} & \textbf{99.35$\pm$0.18} \\
 & E-Doorbell & \textbf{99.45$\pm$0.10} & \textbf{99.40$\pm$0.10} & \textbf{99.40$\pm$0.10} & \textbf{99.40$\pm$0.10} \\
 & Baby monitor & \textbf{84.13$\pm$0.35} & 83.91$\pm$0.34 & 83.96$\pm$0.33 & \textbf{84.06$\pm$0.42} \\
 & 737-sec camera &\textbf{94.84$\pm$0.26} & 94.66$\pm$0.34 & 94.40$\pm$0.33 & \textbf{94.87$\pm$0.70} \\
 & 838-sec camera & \textbf{98.15$\pm$0.19} & \textbf{98.19$\pm$0.19} & \textbf{98.20$\pm$0.14} & \textbf{98.19$\pm$0.14} \\
 & Webcam & \textbf{98.20$\pm$0.12} & \textbf{98.22$\pm$0.10} & \textbf{98.22$\pm$0.09} & \textbf{98.21$\pm$0.09} \\
 & 1002-sec camera & \textbf{93.67$\pm$1.61} & 87.62$\pm$1.60 & 87.75$\pm$1.44 & 87.91$\pm$1.90 \\
\hline
School & & \textbf{11.11$\pm$0.54} & 11.43$\pm$0.77 & 11.32$\pm$0.69 & 11.27$\pm$0.64 \\
\hline
\# Best & & 15 & 4 & 4 & 6 \\
\hline
\end{tabular}
\end{table*}

\subsection{Results}
First, we investigated the zero-shot domain adaptation performance of the proposed method. 
Table \ref{table:t1} shows the average and standard deviation of accuracies with different test domains with MNIST-r and IoT and 
the average and standard deviation of RMSE over 90\%, 85\%, 75\%, and 50\% of all domains for School.
The proposed method obtained the better results than the other methods in all cases. 
For MNIST-r, the proposed method obviously outperformed the others by large margins.
For IoT, the proposed method greatly improved the performance with D-Doorbell and 1002-sec camera
while preserving the competitive performances in the other test domains.
For School, the proposed method statistically outperformed the others.
Overall, the proposed method, which predicts domain-specific models by inferring latent domain vectors from the sets of the feature vectors,
showed good results for all datasets; therefore, 
we found that the proposed method has better zero-shot domain adaptation ability than other existing methods. 

Second, we present insights on the working of the proposed method on the MNIST-r where 
the domains corresponding to image rotations are easy to interpret.
Figure \ref{fig1} represents the posterior distributions of the latent domain vectors estimated by the proposed method,
where there were two dimension of the latent domain vector. 
We found that respective posterior distributions are consistently aligned in order of the rotation angle (domain). 
Especially, the posterior distributions for the unseen domains can also be predicted in order.
For example, when the unseen domain was 30 degrees, the posterior for the 30-degree domain was 
correctly located between the 15 and 45-degree domains. 
Since the proposed method can successfully extract the characteristics of the domains as the latent domain vectors as shown here,
it performed well.

Third, we investigated how the zero-shot domain adaptation performance of the proposed method changed as the number of the dimensions 
of the latent domain vectors $K$ changed.
Figure \ref{fig2} represent the average of the results and their standard error over test domains of each dataset when changing the value of $K$ within $\{2,3,\dots,10 \}$.
Here, Baseline had constant results when the value of $K$ was varied since it does not depend on $K$.
The proposed method performed better than Baseline for all the values of $K$.
Since the proposed method is based on a Bayesian framework,
it is robust to the number of dimensions of the domain vectors $K$.

Lastly, we show how the zero-shot domain adaptation performance is affected by the number of the source domains
on School, which has many domains.
Figure \ref{fig3} represents 
the averages and the standard errors of the RMSE  on School when varying the number of the source domains.
When the number of source domains was large (50\%),
the proposed method and Baseline performed almost the same 
since the amount of training data was sufficient to cover the domain variation.
As the number of source domains decreased,
the proposed method came to perform better than the others.
Since the proposed method is designed to generalize to unseen domains on the basis of a Bayesian framework,
it performed well even when the number of the source domains was small.

\begin{figure*}[t]
  \begin{minipage}{0.245\hsize}
      \centering
      \includegraphics[width=3.85cm]{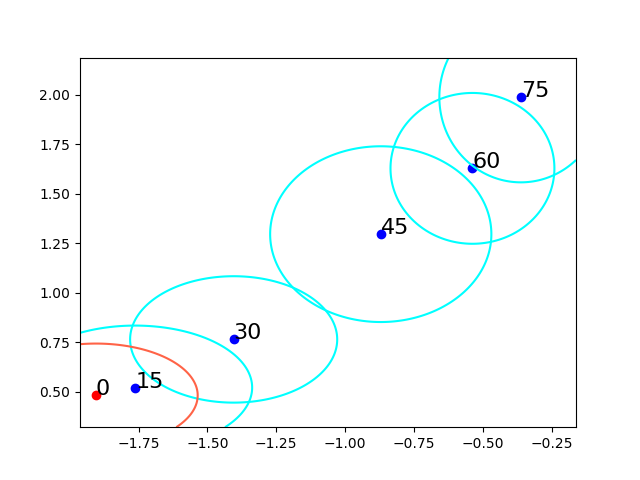}
      \subcaption{0 deg.}
  \end{minipage}
  \begin{minipage}{0.245\hsize}
      \centering
      \includegraphics[width=3.85cm]{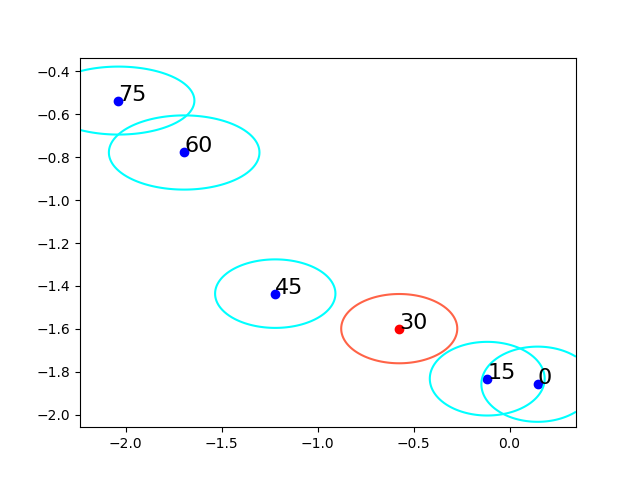}
      \subcaption{30 deg.}
  \end{minipage}
  \begin{minipage}{0.245\hsize}
      \centering
      \includegraphics[width=3.85cm]{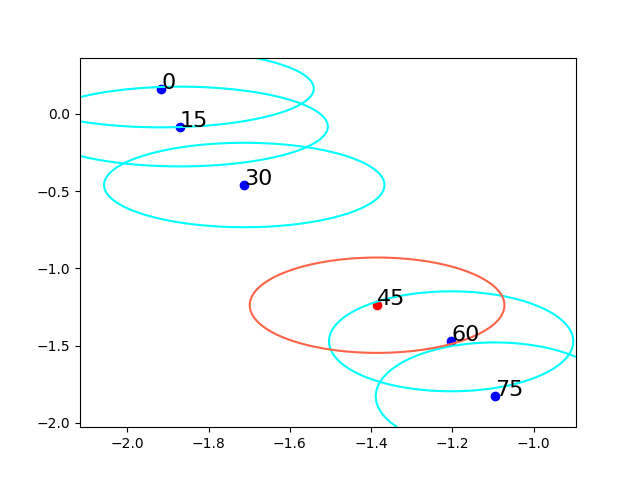}
      \subcaption{45 deg.}
  \end{minipage}
  \begin{minipage}{0.245\hsize}
      \centering
      \includegraphics[width=3.85cm]{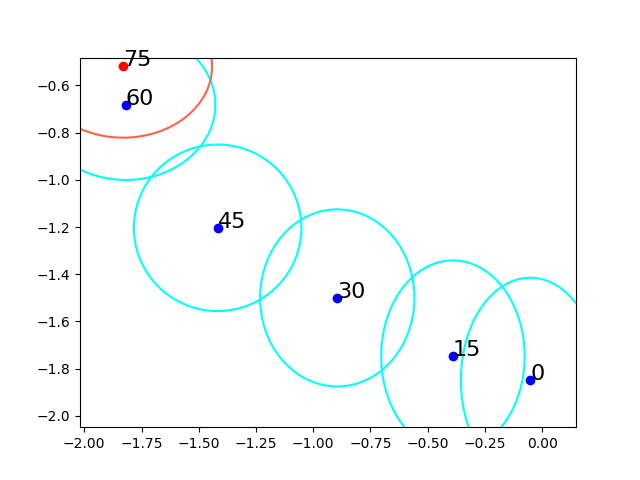}
      \subcaption{75 deg.}
  \end{minipage} 
  \caption{Posterior distributions of the latent domain vectors estimated by the proposed method with MNIST-r.
  Blue and red points represent the mean of the posterior of the latent domain vector 
  for the source domain and the unseen domain, respectively.
  The cyan and red lines represent the contour lines of each posterior distribution.}
  \label{fig1}
\end{figure*}


\begin{figure*}[t]
  \begin{minipage}{0.73\hsize}
  \begin{minipage}{0.326\hsize}
      \centering
      \includegraphics[width=3.3cm]{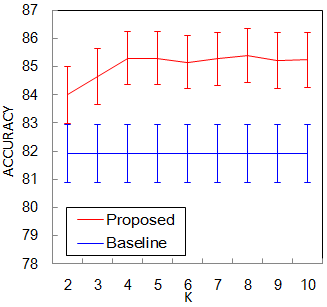}
      \subcaption{MNIST-r}
  \end{minipage}
  \begin{minipage}{0.326\hsize}
      \centering
      \includegraphics[width=3.3cm]{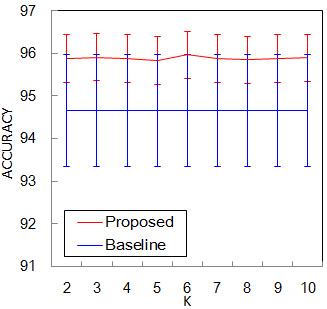}
      \subcaption{IoT}
  \end{minipage}
  \begin{minipage}{0.326\hsize}
      \centering
      \includegraphics[width=3.3cm]{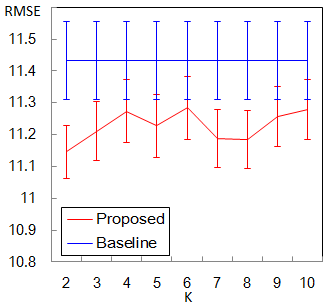}
      \subcaption{School}
  \end{minipage} 
  \caption{Average of the results and their standard error over the test domains of each dataset when the value of $K$ was changed.}
  \label{fig2}
  ~~
  \end{minipage}
  ~
  \begin{minipage}{0.23\hsize}
      \centering
      \includegraphics[width=3.5cm]{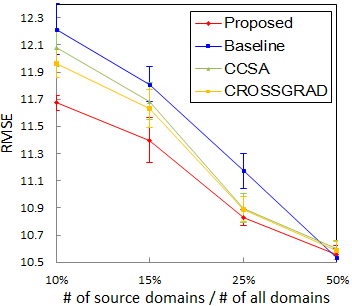}
      \caption{Average RMSE  on School when the number of source domains was varied.}
      \label{fig3}
  \end{minipage}
\end{figure*}

\section{Conclusion}
In this paper, we proposed a method to infer domain-specific models 
such as classifiers for unseen domains without domain semantic descriptors.
To infer domain-specific models without any domain semantic descriptors, 
the proposed method introduces the concept of latent domain vectors,
which are latent representations for domains and are used for inferring the models.
The latent domain vector is estimated from the set of the feature vectors in the corresponding domain.
In experiments using three real-world datasets,
we demonstrated that the proposed method performed
better than the existing zero-shot domain adaptation methods.
For future work, we will incorporate domain semantic descriptors into our framework for learning the latent domain vectors.



\begin{thebibliography}{10}

\bibitem{babar2010proposed}
S.~Babar, P.~Mahalle, A.~Stango, N.~Prasad, and R.~Prasad.
\newblock Proposed security model and threat taxonomy for the internet of
  things (iot).
\newblock In {\em ICNSA}, 2010.

\bibitem{ben2007analysis}
S.~Ben-David, J.~Blitzer, K.~Crammer, and F.~Pereira.
\newblock Analysis of representations for domain adaptation.
\newblock In {\em NIPS}, 2007.

\bibitem{nips2017benaim}
S.~Benaim and L.~Wolf.
\newblock One-sided unsupervised domain mapping.
\newblock In {\em NIPS}, 2017.

\bibitem{blitzer2007}
J.~Blitzer, M.~Dredze, F.~Pereira, et~al.
\newblock Biographies, bollywood, boom-boxes and blenders: domain adaptation
  for sentiment classification.
\newblock In {\em ACL}, 2007.

\bibitem{kon2017cvpr}
K.~Bousmalis, N.~Silberman, D.~Dohan, D.~Erhan, and D.~Krishnan.
\newblock Unsupervised pixel-level domain adaptation with generative
  adversarial networks.
\newblock In {\em CVPR}, 2017.

\bibitem{dai2007}
W.~Dai, Q.~Yang, G.-R. Xue, and Y.~Yu.
\newblock Boosting for transfer learning.
\newblock In {\em ICML}, 2007.

\bibitem{daume2009frustratingly}
H.~Daum{\'e}~III.
\newblock Frustratingly easy domain adaptation.
\newblock {\em ACL}, 2007.

\bibitem{donahue2014decaf}
J.~Donahue, Y.~Jia, O.~Vinyals, J.~Hoffman, N.~Zhang, E.~Tzeng, and T.~Darrell.
\newblock Decaf: a deep convolutional activation feature for generic visual
  recognition.
\newblock In {\em ICML}, 2014.

\bibitem{duan2009domain}
L.~Duan, I.~W. Tsang, D.~Xu, and S.~J. Maybank.
\newblock Domain transfer svm for video concept detection.
\newblock In {\em CVPR}, 2009.

\bibitem{ghifary2015domain}
M.~Ghifary, W.~Bastiaan~Kleijn, M.~Zhang, and D.~Balduzzi.
\newblock Domain generalization for object recognition with multi-task
  autoencoders.
\newblock In {\em ICCV}, 2015.

\bibitem{glorot2011}
X.~Glorot, A.~Bordes, and Y.~Bengio.
\newblock Domain adaptation for large-scale sentiment classification: a deep
  learning approach.
\newblock In {\em ICML}, 2011.

\bibitem{higgins2017darla}
I.~Higgins, A.~Pal, A.~A. Rusu, L.~Matthey, C.~P. Burgess, A.~Pritzel,
  M.~Botvinick, C.~Blundell, and A.~Lerchner.
\newblock Darla: improving zero-shot transfer in reinforcement learning.
\newblock {\em ICML}, 2017.

\bibitem{hoffman2012discovering}
J.~Hoffman, B.~Kulis, T.~Darrell, and K.~Saenko.
\newblock Discovering latent domains for multisource domain adaptation.
\newblock In {\em ECCV}. 2012.

\bibitem{khosla2012undoing}
A.~Khosla, T.~Zhou, T.~Malisiewicz, A.~A. Efros, and A.~Torralba.
\newblock Undoing the damage of dataset bias.
\newblock In {\em ECCV}, 2012.

\bibitem{kingma2014adam}
D.~P. Kingma and J.~Ba.
\newblock Adam: a method for stochastic optimization.
\newblock {\em arXiv}, 2014.

\bibitem{kingma2013auto}
D.~P. Kingma and M.~Welling.
\newblock Auto-encoding variational bayes.
\newblock {\em ICLR}, 2014.

\bibitem{kumagai2016}
A.~Kumagai and T.~Iwata.
\newblock Learning future classifiers without additional data.
\newblock In {\em AAAI}, 2016.

\bibitem{kumagai2017}
A.~Kumagai and T.~Iwata.
\newblock Learning non-linear dynamics of decision boundaries for maintaining
  classification performance.
\newblock In {\em AAAI}, 2017.

\bibitem{lampert2015predicting}
C.~H. Lampert.
\newblock Predicting the future behavior of a time-varying probability
  distribution.
\newblock In {\em CVPR}, 2015.

\bibitem{li2017deeper}
D.~Li, Y.~Yang, Y.-Z. Song, and T.~M. Hospedales.
\newblock Deeper, broader and artier domain generalization.
\newblock In {\em ICCV}, 2017.

\bibitem{li2017learning}
D.~Li, Y.~Yang, Y.-Z. Song, and T.~M. Hospedales.
\newblock Learning to generalize: meta-learning for domain generalization.
\newblock In {\em AAAI}, 2018.

\bibitem{long2016deep}
M.~Long, J.~Wang, and M.~I. Jordan.
\newblock Deep transfer learning with joint adaptation networks.
\newblock {\em ICML}, 2017.

\bibitem{long2016unsupervised}
M.~Long, H.~Zhu, J.~Wang, and M.~I. Jordan.
\newblock Unsupervised domain adaptation with residual transfer networks.
\newblock In {\em NIPS}, 2016.

\bibitem{mancini2018boosting}
M.~Mancini, L.~Porzi, S.~R. Bul{\`o}, B.~Caputo, and E.~Ricci.
\newblock Boosting domain adaptation by discovering latent domains.
\newblock In {\em CVPR}, 2018.

\bibitem{mansour2009domain}
Y.~Mansour, M.~Mohri, and A.~Rostamizadeh.
\newblock Domain adaptation with multiple sources.
\newblock In {\em NIPS}, 2009.

\bibitem{motiian2017unified}
S.~Motiian, M.~Piccirilli, D.~A. Adjeroh, and G.~Doretto.
\newblock Unified deep supervised domain adaptation and generalization.
\newblock In {\em ICCV}, 2017.

\bibitem{muandet2013domain}
K.~Muandet, D.~Balduzzi, and B.~Sch{\"o}lkopf.
\newblock Domain generalization via invariant feature representation.
\newblock In {\em ICML}, 2013.

\bibitem{oh2017zero}
J.~Oh, S.~Singh, H.~Lee, and P.~Kohli.
\newblock Zero-shot task generalization with multi-task deep reinforcement
  learning.
\newblock {\em ICML}, 2017.

\bibitem{qiu2012domain}
Q.~Qiu, V.~M. Patel, P.~Turaga, and R.~Chellappa.
\newblock Domain adaptive dictionary learning.
\newblock In {\em ECCV}, 2012.

\bibitem{ruder2017overview}
S.~Ruder.
\newblock An overview of multi-task learning in deep neural networks.
\newblock {\em arXiv}, 2017.

\bibitem{saenko2010adapting}
K.~Saenko, B.~Kulis, M.~Fritz, and T.~Darrell.
\newblock Adapting visual category models to new domains.
\newblock {\em ECCV}, 2010.

\bibitem{salimans2016improved}
T.~Salimans, I.~Goodfellow, W.~Zaremba, V.~Cheung, A.~Radford, and X.~Chen.
\newblock Improved techniques for training gans.
\newblock In {\em NIPS}, 2016.

\bibitem{schuster2010speech}
M.~Schuster.
\newblock Speech recognition for mobile devices at google.
\newblock In {\em PRICAI}, 2010.

\bibitem{shankar2018generalizing}
S.~Shankar, V.~Piratla, S.~Chakrabarti, S.~Chaudhuri, P.~Jyothi, and
  S.~Sarawagi.
\newblock Generalizing across domains via cross-gradient training.
\newblock In {\em ICLR}, 2018.

\bibitem{tokui2015chainer}
S.~Tokui, K.~Oono, S.~Hido, and J.~Clayton.
\newblock Chainer: a next-generation open source framework for deep learning.
\newblock In {\em Proceedings of workshop on machine learning systems
  (LearningSys) in NIPS}, 2015.

\bibitem{2011unbiased}
A.~Torralba and A.~A. Efros.
\newblock Unbiased look at dataset bias.
\newblock In {\em CVPR}, 2011.

\bibitem{tzeng2014deep}
E.~Tzeng, J.~Hoffman, N.~Zhang, K.~Saenko, and T.~Darrell.
\newblock Deep domain confusion: maximizing for domain invariance.
\newblock {\em arXiv}, 2014.

\bibitem{xian2017zero}
Y.~Xian, C.~H. Lampert, B.~Schiele, and Z.~Akata.
\newblock Zero-shot learning-a comprehensive evaluation of the good, the bad
  and the ugly.
\newblock {\em arXiv}, 2017.

\bibitem{xiao2017building}
X.~Xiao, L.~Jin, Y.~Yang, W.~Yang, J.~Sun, and T.~Chang.
\newblock Building fast and compact convolutional neural networks for offline
  handwritten chinese character recognition.
\newblock {\em Pattern Recognition}, 72:72--81, 2017.

\bibitem{xiong2014latent}
C.~Xiong, S.~McCloskey, S.-H. Hsieh, and J.~J. Corso.
\newblock Latent domains modeling for visual domain adaptation.
\newblock In {\em AAAI}, 2014.

\bibitem{yang2015zero}
Y.~Yang and T.~Hospedales.
\newblock Zero-shot domain adaptation via kernel regression on the
  grassmannian.
\newblock {\em BMVC workshop on Differential Geometry in Computer Vision},
  2015.

\bibitem{yang2014unified}
Y.~Yang and T.~M. Hospedales.
\newblock A unified perspective on multi-domain and multi-task learning.
\newblock In {\em ICLR}, 2015.

\bibitem{NIPS2017_6931}
M.~Zaheer, S.~Kottur, S.~Ravanbakhsh, B.~Poczos, R.~R. Salakhutdinov, and A.~J.
  Smola.
\newblock Deep sets.
\newblock In {\em NIPS}, 2017.

\bibitem{zhang2015multi}
K.~Zhang, M.~Gong, and B.~Sch{\"o}lkopf.
\newblock Multi-source domain adaptation: a causal view.
\newblock In {\em AAAI}, 2015.

\end{thebibliography}

\end{document}